\documentclass[journal]{IEEEtran}

\usepackage[english]{babel}
\usepackage[utf8x]{inputenc}
\usepackage[T1]{fontenc}
\usepackage{cite}

\usepackage{amsmath,amssymb,mathrsfs,bbm}
\usepackage{graphicx,balance}
\usepackage[colorinlistoftodos]{todonotes}
\usepackage[colorlinks=true, allcolors=blue]{hyperref}
\usepackage{multirow}
\usepackage[lined,linesnumbered,ruled]{algorithm2e}

\usepackage{psfrag,epsfig,graphics}
\usepackage{amsmath,amsthm,amssymb,multirow}
\usepackage{mathbbol}
\usepackage{amssymb}

\DeclareSymbolFontAlphabet{\amsmathbb}{AMSb}%

\newcommand{\cp}[1]{\ifmmode {\mathcal{#1}}\else ${\mathcal{#1}}$\fi}
	
\newcommand{\bA}{\boldsymbol{A}}

\newcommand{\bI}{\boldsymbol{I}}

\newcommand{\bM}{\boldsymbol{M}}

\newcommand{\bX}{\boldsymbol{X}}
\newcommand{\bY}{\boldsymbol{Y}}

\newcommand{\ba}{\boldsymbol{a}}

\newcommand{\bm}{\boldsymbol{m}}

\newcommand{\be}{\boldsymbol{e}}

\newcommand{\by}{\boldsymbol{y}}

\newcommand{\bx}{\boldsymbol{x}}

\newcommand{\bz}{\boldsymbol{z}}

\newcommand{\calD}{\mathcal{D}}

\newcommand{\calG}{\mathcal{G}}
\newcommand{\calM}{\mathcal{M}}

\newcommand{\cb}[1]{\boldsymbol{#1}}

\newcommand{\Ex}{\operatorname{E}}

\usepackage{color}  

\definecolor{darkgreen}{rgb}{0., 0.4, 0.}

\title{Deep Generative Models for Library Augmentation in Multiple Endmember Spectral Mixture Analysis}

\author{Ricardo~Augusto~Borsoi, Tales Imbiriba,~\IEEEmembership{Member,~IEEE,} Jos\'e~Carlos~Moreira~Bermudez,~\IEEEmembership{Senior~Member,~IEEE}, C\'edric Richard,~\IEEEmembership{Senior~Member,~IEEE}
\thanks{This work has been supported by the National Council for Scientific and Technological Development (CNPq) under grants 304250/2017-1, 409044/2018-0, 141271/2017-5 and 204991/2018-8, and by the Foundation for Research Support of the State of Rio Grande do Sul (FAPERGS) under grant 19/2551-0001844-4.}
\thanks{R.A. Borsoi is with the Department of Electrical Engineering, Federal University of Santa Catarina (DEE--UFSC), Florian\'opolis, SC, Brazil, and with the Lagrange Laboratory (CNRS, OCA), Universit\'e  C\^ote  d'Azur, Nice, France. e-mail: \mbox{raborsoi@gmail.com}.}
\thanks{T. Imbiriba is with the ECE department of the Northeastern University, Boston, MA, USA. e-mail: \mbox{talesim@gmail.com}.}
\thanks{J.C.M. Bermudez is with the DEE--UFSC, Florian\'opolis, SC, Brazil, and with the Graduate Program on Electronic Engineering and Computing, Catholic University of Pelotas (UCPel) Pelotas, Brazil. e-mail: \mbox{j.bermudez@ieee.org}.}
\thanks{C. Richard is with the  Lagrange Laboratory (CNRS, OCA), Universit\'e  C\^ote  d'Azur,  Nice, France. e-mail: \mbox{cedric.richard@unice.fr}.}
}

\begin{document}
\maketitle

\begin{abstract}

Multiple Endmember Spectral Mixture Analysis (MESMA) is one of the leading approaches to perform spectral unmixing (SU) considering variability of the endmembers (EMs). It represents each EM in the image using libraries of spectral signatures acquired \textit{a priori}. However, existing spectral libraries are often small and unable to properly capture the variability of each EM in practical scenes, which compromises the performance of MESMA. In this paper, we propose a library augmentation strategy to increase the diversity of existing spectral libraries, thus improving their ability to represent the materials in real images. First, we leverage the power of deep generative models to learn the statistical distribution of the EMs based on the spectral signatures available in the existing libraries. Afterwards, new samples can be drawn from the learned EM distributions and used to augment the spectral libraries, improving the overall quality of the SU process. Experimental results using synthetic and real data attest the superior performance of the proposed method even under library mismatch conditions.
\end{abstract}


\section{Introduction}


Spectral Unmixing (SU) aims at extracting the spectral signatures of materials present in the hyperspectral images (HI) of a scene, which are called endmembers (EMs), as well as the proportion to which they contribute to each HI pixel~\cite{Keshava:2002p5667}. The SU problem can be solved using algorithms that are either supervised, where the EMs are known a priori, or unsupervised, where the EMs are estimated from the HI~\cite{ammanouil2014blind}.
The most popular model to describe the interaction between light and the targets is the \textit{Linear Mixing Model} (LMM), which represents the reflectance at each pixel as a convex combination of the spectral signatures of the EMs~\cite{Keshava:2002p5667}.
However, the LMM fails to represent important nonideal effects observed in practice, such as nonlinear interactions between light and the materials~\cite{heylen2014review,Imbiriba2016_tip,Imbiriba2017_bs_tip} and variations of the EM spectra along the scene~\cite{somers2011variabilityReview,imbiriba2018glmm}.

EM variability is an important effect originating from environmental, illumination or atmospheric changes which may lead to significant estimation errors in SU~\cite{somers2011variabilityReview}.
The most prominent approach to deal with EM variability in SU consists in modeling EMs as sets of spectral signatures, also called spectral libraries~\cite{somers2011variabilityReview}.
The spectral signatures in each library are variants of a material produced under different acquisition conditions or physico-chemical compositions. They are usually acquired a priori through laboratory or \textit{in situ} measurements.
The SU problem then becomes equivalent to selecting a subset of signatures in the libraries that can best represent the observed HI under the LMM. The methods that attempt to solve this problem can be roughly divided between sparse SU~\cite{iordache2012sunsal_TV,Borsoi_multiscale_lgrs_2018} and \emph{Multiple Endmember Spectral Mixture Analysis} (MESMA)~\cite{roberts1998originalMESMA} algorithms.
The MESMA algorithm is widely used due to its simplicity and interpretability, and has been widely employed in practice~\cite{somers2011variabilityReview}.
However, the quality of the MESMA results is strongly dependent on how well the spectral libraries represent the EM signatures actually present in the scene.
This is a problem since spectral libraries are usually not acquired under the same conditions as the observed HI, since \textit{in situ} measurements can be costly or impractical. Furthermore, most existing spectral libraries only have very few signatures of each material, and might not adequately capture spectral variability occurring in the scene.


One approach to alleviate this problem consists of generating multiple synthetic samples of an endmember using a physical model (radiative transfer function -- RTF) describing the variability of the spectra as a function of atmospheric or biophysical parameters~\cite{somers2011variabilityReview}, such as e.g. the PROSPECT of Hapke models~\cite{jacquemoud1990PROSPECTmodelLeaf,Hapke1981} for vegetation or mineral spectra. These additional signatures are then included in the library to augment it before performing SU.
The use of RTFs to generate spectral libraries has great potential since it can represent spectral variability caused by different effects which are unlike to be captured by laboratory or field measurements~\cite{peddle1999canopyModelLibraryBorealRTF,dennison2006wildfireTemperatureMESMA_RTF,somers2009modelSoilSpectraMoisture}.
However, physics-based models require accurate knowledge of the physical process governing the observation of the materials spectra by the sensor, which is hard to obtain in practice. This limits the practical interest of these methods.

Recently, deep generative models (DGMs) have seen remarkable advances in the form of variational autoencoders (VAEs) and generative adversarial networks (GANs)~\cite{kingma2013AEC_varBayes, goodfellow2014GANs}. This have made it possible to learn the distribution of complex data (e.g., natural images) efficiently, and from a limited amount of samples~\cite{antoniou2017dataAugmentationGANs}.
DGMs have been considered for data augmentation in few-sample settings for image classification problems~\cite{antoniou2017dataAugmentationGANs}. 
A recent work proposed to use DGMs learned from observed HIs in order to parametrize the variable EM spectra in the optimization step of a matrix factorization-based blind SU problem, where the EMs are estimated from the HI~\cite{borsoi2019deep}. This showed that using generative neural networks is a promising approach to represent the \mbox{EMs in SU.}


In this paper, we propose a spectral library augmentation method for MESMA-based algorithms by leveraging the power of DGMs to represent the EMs. 
%
The main contribution of the proposed method is that it works blindly, what allows for augmentation of the spectral libraries used with MESMA even when RTFs or physical models are unknown.
The overall strategy can be divided in three steps. First we learn the statistical distribution of each EM in the scene using the spectral signatures contained in the existing spectral library and a DGM. Then, we sample new spectral signatures using the DGMs and augment their respective spectral libraries. Finally, we unmix the observed HI using MESMA and the augmented library. Simulations with synthetic and real data show a substantial accuracy gain in abundance estimation when comparing the proposed method with \mbox{competing strategies.}
\section{Spectral unmixing with MESMA}\label{sec:MESMA}

Most MESMA algorithms consider the LMM as their central building block. The LMM assumes that each $L$-band pixel $\by_n\in\amsmathbb{R}^L$, $n=1,\ldots, N$, of a $N$-pixel HI, can be modeled as:
\begin{align} \label{eq:LMM}
    &\by_n = \bM \ba_n + \be_n, 
    \,\,\,\, \text{s. t. }\,\,\,\,\cb{1}^\top\ba_n = 1 \text{ and } \ba_n \geq \cb{0} 
\end{align}
where $\bM \in \amsmathbb{R}^{L\times P}$ is a matrix whose columns are the $P$ EM spectral signatures $\bm_k$, $\ba_n$ is the abundance vector and $\be_n$ is an additive noise term.
Differently from most LMM-based SU methodologies, which assume a unique EM for each material in the scene, MESMA considers multiple spectra libraries, or bundles, one for each endmember, and performs a search for the best fitting model within all possible combinations of endmembers.
Thus, assuming prior knowledge of spectral bundles for each EM in the scene, the set $\calM$ of endmember matrices that can be drawn from the library can be defined as
\begin{align}
	\calM {}={} \Big\{\big[\bm_1,\ldots,\bm_P\big]\,:\,
	\bm_k\in\mathcal{M}_k,\,k=1,\ldots,P\Big\}
\end{align}
where $\calM_k=\{\bm_{k,1},\ldots,\bm_{k,C_k}\}$, $\bm_{k,j}\in\amsmathbb{R}^L$ is a set of $C_k$ spectral signatures of the $k^{\rm th}$ material. The MESMA SU problem can be formulated as 
\begin{align} \label{eq:mesma}
\begin{split}
	\min_{\bM\in\calM,\,\ba_n} 
    \big\|\by_n - \bM\ba_n\big\|_2^2 \,\,\,\,
    \text{s. t. } \,\,  \ba_{n} \geq\cb{0}, \, \cb{1}^\top\ba_{n} = 1.
\end{split}
\end{align}
Although the MESMA algorithm has shown excellent performance when dealing with spectral variability in many practical scenarios, its performance is strongly effected by the quality of the spectral library~$\calM$~\cite{somers2011variabilityReview}. In order for MESMA to perform well, the library must be representative of the spectral library observed in a given scene.
Previous works tried to address this issue by augmenting the spectral libraries using physics-based models that describe well the variability of the endmembers. See, e.g., the PROSPECT or Hapke models~\cite{jacquemoud1990PROSPECTmodelLeaf,Hapke1981}. 

However, a major drawback of physics-based models is the requirement of accurate knowledge of the physical process governing the observation of the materials spectra by the sensor. This detailed information is rarely available in practice, which limits the applicability of these methods.
In the following, we will present a new approach for spectral library augmentation that is based on deep generative models such as VAEs and GANs. These approaches allows one to learn the statistical distribution of the endmembers from very few training samples, making it effective in practical scenarios.

\begin{figure}[!t]
    \vspace{-1ex}
    \centering
    \includegraphics[width=\linewidth]{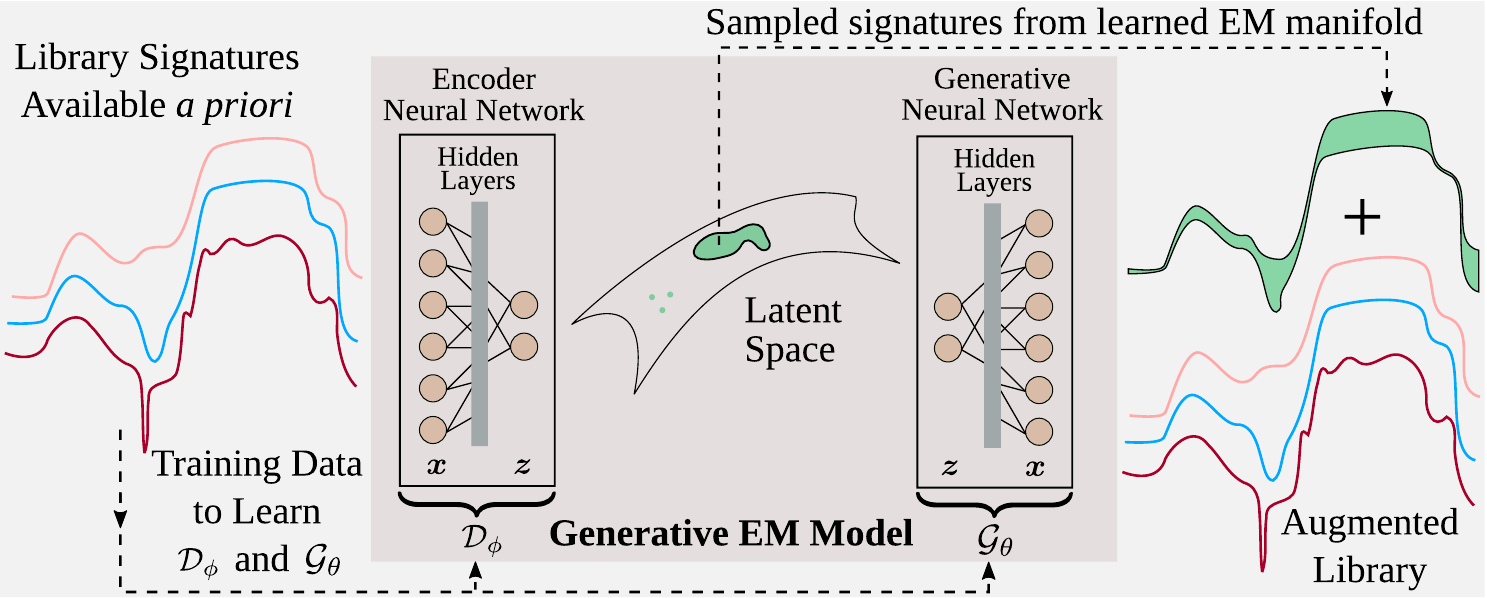}
    \vspace{-0.65cm}
    \caption{Outline of the proposed approach: deep generative models are used to approximate the distribution of spectra belonging to a library. Then, new spectral samples (right) can be obtained by propagating samples drawn from the EM submanifold through $\calG_\theta$ and used to augment the spectral library.}
    \label{fig:method_illustration}
    \vspace{-2ex}
\end{figure}

\section{Library Augmentation with DGMs}\label{sec:LibraryAug}

Physics-based models describing the variations of the spectral signatures in a scene reveal an important characteristic of spectral variability: that EM spectra usually lies on a low-dimensional submanifold of the high-dimensional spectral space~$\amsmathbb{R}^L$.
This assumption is in agreement with most physical models, such as the PROSPECT or Hapke's~\cite{jacquemoud1990PROSPECTmodelLeaf,Hapke1981}, which represent the spectral signature of the materials as a function of only a small number of photometric or chemical properties.

Instead of employing physics-based models, we propose in this paper to augment the spectral libraries by using deep generative models. Generative models aim to estimate the probability distribution $p(X)$ of a random variable $X\in\amsmathbb{R}^L$ based on a set of $N_x$ observations $\bx_i$. Then, they allow one to generate new samples that look similar to new realizations of~$X$. Such models have shown good performance at representing endmember spectra in blind unmixing applications~\cite{borsoi2019deep}. Here we propose to use the signatures in existing spectral libraries to learn the generative models describing the distributions of EM spectra. Then, to enhance the ability of MESMA to adapt to a wider range of spectral variability, we augment the libraries by sampling from the estimated distributions. An illustrative outline of this strategy is shown in Fig.~\ref{fig:method_illustration}.

%
%

Even though the spectral dimensionality $L$ is high compared to the small number of signatures often found in typical spectral libraries (making this problem very hard in general~\cite{neal2001annealedDensityEstimation,arjovsky2017wassersteinGANs}), the low-dimensinality of the manifolds to which the EM spectra is confined, allied with recent advances in generative models, have made this problem tractable. This framework has shown success in capturing the distribution of complex data such as natural images from very few training samples~\cite{antoniou2017dataAugmentationGANs}, which illustrates its appropriateness for our application.
%


\paragraph*{\textbf{Deep generative models}}
A convenient way to estimate the PDF $p(X)$ of a random variable $X$ that lies on a low-dimensional submanifold of $\amsmathbb{R}^L$ is to define a new random variable $\amsmathbb{R}^K\ni Z\sim p(Z)$, with $K\ll L$ and a known distribution~$p(Z)$, and a parametric function (e.g. a neural network) $\mathcal{G}_{\theta}$ which maps $Z\mapsto \widehat{X}\in\amsmathbb{R}^L$ such that the distribution of the transformed random variable $\widehat{X}=\mathcal{G}_{\theta}(Z)$ is very close to $p(X)$.
This allows us to generate new samples from $\widehat{X}$ by first sampling from $Z\sim p(Z)$ and then applying the function $\mathcal{G}_{\theta}(Z)$.
Although estimating $\mathcal{G}_{\theta}$ to fulfill this objective might seem difficult, recent advances in generative modeling such as VAEs~\cite{kingma2013AEC_varBayes} and GANs~\cite{goodfellow2014GANs} have shown excellent performance for modeling complex distributions (e.g., of natural images) using only a limited amount of samples~\cite{antoniou2017dataAugmentationGANs}.

VAEs address this problem by maximizing a lower bound on the log-likelihood of $p(X)$~\cite{kingma2013AEC_varBayes}:
\begin{align} 
    \log p(X) \! \geq &
    \Ex_{q_\phi(Z|X)} \! \big\{ \! \log p(X|Z) \big\}
    \!-\! KL\big(q_\phi(Z|X)\|p(Z)\big) \,,
    \nonumber
\end{align}
where the function~$\mathcal{G}_\theta$ is represented by $p(X|Z)$, $KL(\cdot\|\cdot)$ is the Kullback-Leibler divergence between two distributions, $\Ex_{\varsigma}\{\cdot\}$ is the expected value operator with respect to the distribution  $\varsigma$ and $q_\phi(Z|X)$ is a variational approximation to the intractable posterior $p(Z|X)$, which is also represented through another parametric function~$\mathcal{D}_{\phi}:\amsmathbb{R}^L\to\amsmathbb{R}^K$.

Differently, GANs attempt to learn the distribution $p(X)$ by seeking for the Nash equilibrium of a two-player adversarial game~\cite{goodfellow2014GANs} between the generator network $\mathcal{G}_\theta$ and a discriminator network $\mathcal{C}_\phi$, which predicts the probability of a sample $\bx_i$ coming from the true distribution~$p(X)$ instead of being generated through $\mathcal{G}_\theta$. The generator $\mathcal{G}_\theta$ is trained to maximize the probability of the discriminator making a mistake, which is formulated as the following minimax optimization problem:
\begin{align} 
    \min_{\mathcal{G}_\theta} \, \max_{\mathcal{C}_\phi} \,\,\,  &
    \Ex_{p(X)}\big\{\!\log \mathcal{C}_{\phi}(X) \big\}
    + \Ex_{p(Z)}\big\{(1-\mathcal{C}_{\phi}(\mathcal{G}_\theta(Z)))\big\}
    \,.
    \nonumber
\end{align}
Although GANs are more flexible and have shown better results when modeling complex distributions, they are also much harder to train~\cite{arjovsky2017wassersteinGANs}. This motivated us to use VAEs in this work due their more stable training procedure.
Future works will consider the use of GANs.


\paragraph*{\textbf{Library augmentation}}
Consider a small spectral library $\calM$ known a priori containing a set of spectral signatures $\calM_i$ for each material $i=1,\ldots,P$. Each signature $\bm_{i,j}\in\calM_i$, $j=1,\ldots, C_k$, can be viewed as a sample drawn from the statistical distribution of the $i^{\rm th}$ EM spectra.
Thus, these libraries can be employed as training data to learn a set of generative models $\calG_{\theta_i}$ that represents the probability distribution function $p_i(M)$ of each EM $i=1,\ldots,P$ using a VAE~\cite{kingma2013AEC_varBayes}.

Given the learned generative models $\calG_{\theta_i}$, we can then generate new spectral signatures from each EM class by sampling from the distribution of $\calG_{\theta_i}(Z)$, where $Z\sim\mathcal{N}(0,\bI_K)$. These new signatures can then be used to augment into the original library $\calM$, yielding a new spectral library $\widetilde{\calM}$ which is more comprehensive and better accounts for different spectral variations of each material.
Finally, the MESMA algorithm can be applied to unmix each image pixel $\by_n$ using the augmented library $\widetilde{\calM}$. This procedure is described in detail in Algorithm~\ref{alg:global}, where the spectral library is augmented by adding $N_s$ samples to each EM set.
Note that although this increases the complexity of SU with MESMA, approximate strategies can be used to obtain an efficient solution when the augmented library has many signatures~\cite{heylen2016alternatingAngleMinimization}.

\begin{algorithm} [t]
\footnotesize
\SetKwInOut{Input}{Input}
\SetKwInOut{Output}{Output}
\caption{MESMA with spectral library augmentation~\label{alg:global}}
\Input{$\bY$, $\mathcal{M}_i$, $i=1,\ldots,P$ and $N_s$.}
\For{$i=1,\ldots,P$}{
Set $\widetilde{\calM}_i=\calM_i$ and train a DGM $\calG_{\theta_i}$ using the samples in $\calM_i$ \;
\For{$j=1,\ldots,N_s$}{
Sample $\bz\sim\mathcal{N}(\cb{0},\bI)$ and compute $\widehat{\bm}=\calG_{{\theta}_i}(\bz)$\;
$\widetilde{\calM}_i {}\leftarrow{} \widetilde{\calM}_i \bigcup \big\{\widehat{\bm}\big\}$ \;
}}
Set $\widetilde{\calM}{}={}\big\{[\bm_1,\ldots,\bm_P]\,:\,\bm_k\in\widetilde{\calM}_i,\,i=1,\ldots,P\big\}$ \;
Run MESMA with the augmented library $\widetilde{\calM}$ to compute~$\,\widehat{\!\bA}$ \;
\KwRet $\,\widehat{\!\bA}$, $\widetilde{\calM}$ \;
\end{algorithm}

\begin{table}[h]
\vspace{-2ex}
\footnotesize
\centering
\renewcommand{\arraystretch}{1.15}
\caption{Encoder and Decoder network architectures.}
\vspace{-0.2cm}
\begin{tabular}{c|ccc}
    \hline
    &Layer       & Activation   & Number of units \\ \hline 
    \multirow{4}{*}{$\mathcal{D}_{\phi}$} 
    &Input        &  ---  & $L$ \\
    &Hidden \# 1 & ReLU  & $\lceil 1.2\times L \rceil + 5$ \\
    &Hidden \# 2 & ReLU  & $\max\big\{\lceil L/4 \rceil,\, K+2\big\} + 3$ \\
    &Hidden \# 3 & ReLU  & $\max\big\{\lceil L/10 \rceil,\, K+1\big\}$ \\[0.05cm]
    \hline 
    \multirow{4}{*}{$\mathcal{G}_{\theta}$} 
    &Hidden \# 1  & ReLU & $\max\big\{\lceil L/10 \rceil,\, K+1\big\}$ \\
    &Hidden \# 2  & ReLU & $\max\big\{\lceil L/4 \rceil,\, K+2\big\} + 3$ \\
    &Hidden \# 3  & ReLU & $\lceil 1.2\times L \rceil + 5$  \\
    &Output       & Sigmoid & $L$ \\
    \hline 
\end{tabular}
\label{tab:net_architecture}
\vspace{-1.5ex}
\end{table}


\paragraph*{\textbf{Network architecture}}
To learn the generative models $\calG_{\theta_p}$, we used a VAE~\cite{kingma2013AEC_varBayes} due to its stable training~\cite{arjovsky2017wassersteinGANs} and because it behaved well with small spectral libraries.
The network architectures for $\calG_{\theta_p}$ and $\calD_{\phi_p}$ and the dimension of the latent spaces were selected as in~\cite{borsoi2019deep} since they resulted in a good experimental performance and showed sufficient capacity to capture the spectral variability of a given library. The network architectures are shown in Table~\ref{tab:net_architecture} and the latent spaces dimension was set to $K=2$.
Finally, the network training was performed with the Adam optimizer~\cite{kingma2014adam} in TensorFlow for 50 epochs.

\vspace{-0.5ex}

\section{Experimental Results} \label{sec:results}

In this section, simulation results using both synthetic and real data illustrate the performance of the proposed method. We compare the performance of MESMA using the augmented library with that of the traditional MESMA algorithm. We also present results obtained with the fully constrained least squares (FCLS) and the the GLMM~\cite{imbiriba2018glmm}, which estimate the endmembers from the observed HI (without using a spectral library).
The VCA algorithm~\cite{Nascimento2005} was used to extract EMs used by the FCLS and GLMM methods.
The performances were evaluated using the Root Mean Squared Error (RMSE) between the estimated abundance maps ($\text{RMSE}_{\bA}$) and between the reconstructed images ($\text{RMSE}_{\bY}$). The RMSE between two matrices is defined as $\text{RMSE}_{\bX} = \sqrt{\|\bX-\bX^*\|^2_F\,/\,N_{\bX}}$,
where $N_{\bX}$ denotes the number of elements in~$\bX$.




\begin{table} 
\footnotesize
\caption{Simulations with synthetic and real data (values $\times10^3$).}
\centering
\renewcommand{\arraystretch}{1.2}
\vspace{-0.2cm}
\resizebox{\linewidth}{!}{%
\begin{tabular}{l|cc|c|c}
\hline
&\multicolumn{2}{|c}{Synthetic HI}  &\multicolumn{1}{|c}{Alunite Hill} &\multicolumn{1}{|c}{Gulfport}\\
\hline
& $\text{RMSE}_{\bA}$ & $\text{RMSE}_{\bY}$ & $\text{RMSE}_{\bY}$ & $\text{RMSE}_{\bY}$ \\ \hline
FCLS     & $50.0\pm32.2$  & $0.73\pm0.87$  & $0.47\pm0.60$      & $1.00\pm2.06$ \\
GLMM     & $45.3\pm31.2$  & $0.30\pm0.22$  & $0.001\pm0.002$    & $0.002\pm0.003$ \\
MESMA    & $18.2\pm13.7$  & $0.41\pm0.45$  & $19.2\pm14.0$      & $1.31\pm2.02$ \\
Proposed & $15.3\pm11.0$  & $0.26\pm0.25$  & $18.4\pm12.8$      & $1.16\pm1.86$ \\
\hline
\end{tabular}}

\vspace{0.075cm}

\resizebox{\linewidth}{!}{%
\begin{tabular}{l|ccccccc}
\hline
\multicolumn{8}{c}{$\text{RMSE}_{\bA}$ of Algorithm~\ref{alg:global} as a function of $N_s$} \\
\hline
$N_s$ & 0 & 1 & 2 & 3 & 4 & 5 & 6\\ \hline
$\text{RMSE}_{\bA}$ & 18.18 & 16.23 & 15.65 & 15.34 & 15.21 & 15.09 & 15.01\\
\hline
\end{tabular}}
\label{tab:results_synthData}
\vspace{-3ex}
\end{table}



\begin{figure}[b]
    \vspace{-2ex}
    \centering
    \begin{minipage}{0.49\linewidth}
    \centering
    \raisebox{-0.5\height}{\includegraphics[width=0.85\linewidth]{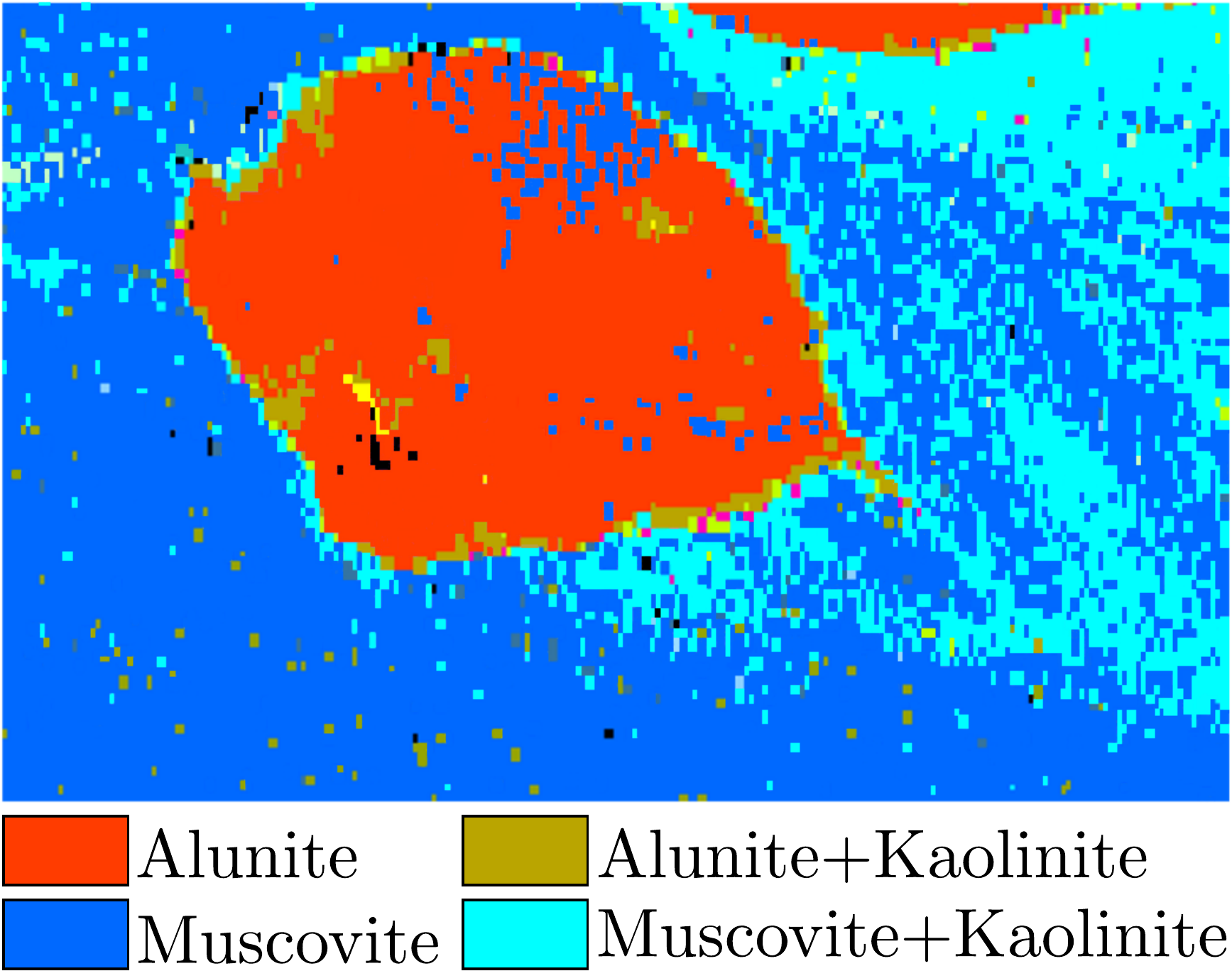}}
    \end{minipage}
    \begin{minipage}{0.49\linewidth}
    \centering
    \raisebox{-0.5\height}{\includegraphics[width=0.85\linewidth]{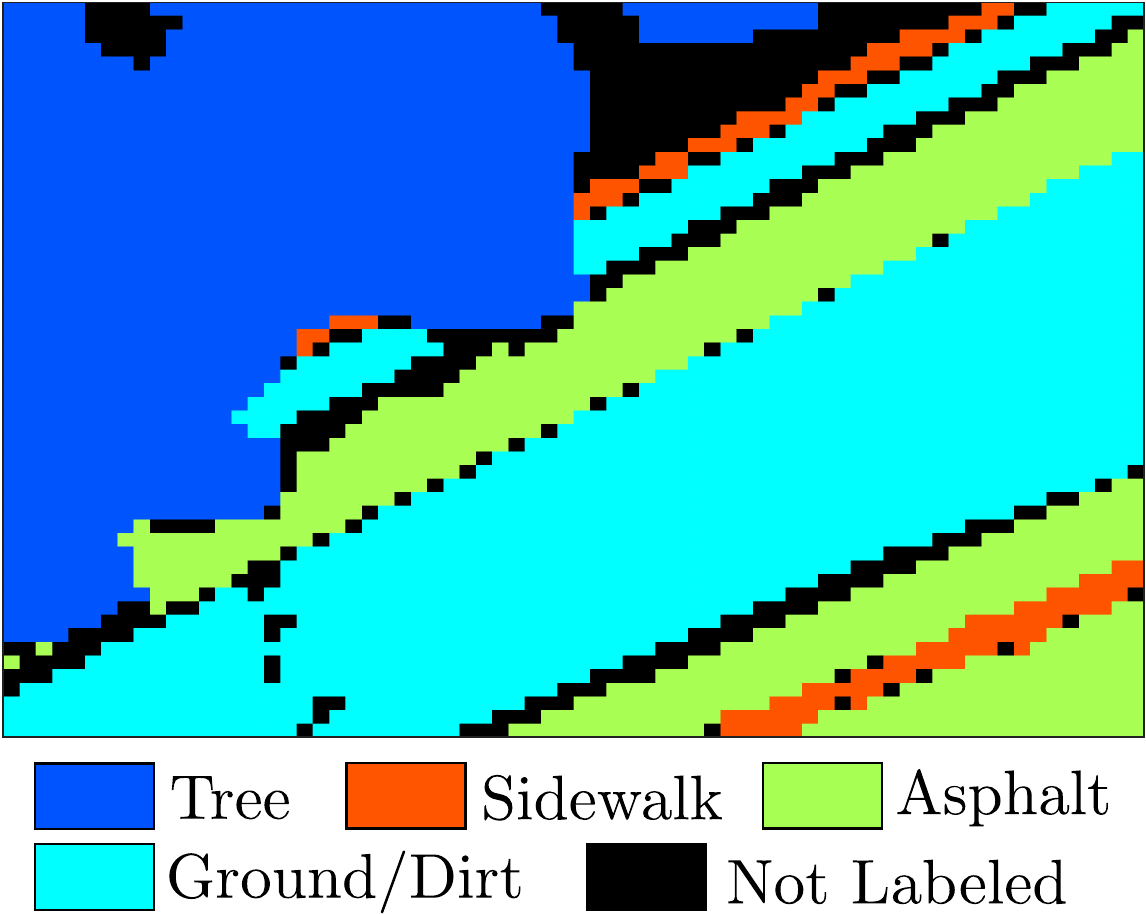}}
    \end{minipage}
    \vspace{-0.3cm}
    \caption{``Ground truth'' for the Alunite Hill (left) and Gulfport (right) HIs.}
    \label{fig:aluniteHillGT}
\end{figure}



\paragraph*{\textbf{Synthetic data with library mismatch}}
In this example, we evaluate the performance of the proposed approach quantitatively using a synthetic data set with three endmembers and $L=198$ spectral bands. 
The goal is to simulate a typical library mismatch scenario often found when considering library-based unmixing~\cite{iordache2012sunsal_TV}.
To generate and process this dataset, we first obtained two disjoint sets of endmember spectra $\mathcal{M}_{i}^1$ and $\mathcal{M}_i^2$, with $\mathcal{M}_{i}^1\cap\mathcal{M}_{i}^2=\varnothing$, $i\in\{1,2,3\}$ by manually extracting pure pixels of soil, vegetation and water from a real hyperspectral scene (the Jasper Ridge HI~\cite{imbiriba2018LRTM}).
%
The sets $\mathcal{M}_{i}^1$ contained 20 signatures each and were used to compose the synthetic pixel spectra $\by_n$, while each of the sets $\mathcal{M}_i^2$ contained 14 signatures that were employed to construct the spectral libraries used by MESMA to perform SU.
We simulated a library mismatch by applying a random affine transformation (a gain and an additive scaling in the intervals $[0.75,1.25]$ and $[-0.15,0.15]$, respectively) to each element of $\mathcal{M}_{i}^1$, $i\in\{1,2,3\}$.
To generate each pixel, we used the LMM considering abundance fractions $\ba_n$ sampled from a Dirichlet distribution with concentration parameters selected such as to have a heavily mixed data in order to evaluate the methods in a challenging scenario, and pixel-dependent endmember matrices obtained by randomly (uniformly) selecting one spectral signature from each of the sets $\mathcal{M}_{i}^1$, $i\in\{1,2,3\}$. White Gaussian noise with a signal to noise ratio (SNR) of 30dB was added to the data.

The final library $\mathcal{M}$ available for the MESMA-based methods was created by sampling five signatures at random of each material from $\mathcal{M}_{i}^2$, and no other preprocessing or adequacy strategy was used to mitigate mismatch between the available library and the true endmembers used to construct the scene. Only the spectra in $\mathcal{M}$ was used to learn the DGMs, and $N_s=3$ additional signatures were sampled for each material.
Finally, in order to provide a proper statistical evaluation, this whole procedure was repeated for $10^{4}$ Monte Carlo realizations. The mean values and standard deviations are shown in Table~\ref{tab:results_synthData}.
It can be seen that despite only a small number of signatures being available to train the DGMs, the proposed strategy provided a substantial (16\%) improvement in the abundance estimation RMSE when compared to the MESMA algorithm. This shows that with a careful selection of the neural network architecture, the proposed method can work even under such challenging conditions. When compared with the other methods the proposed solution improvement is even more significant obtaining gains of 70\% (FCLS) and 67\% (GLMM). These experiments show that the proposed data augmentation strategy can lead to significant performance gains when compared to the plain MESMA algorithm.

To investigate the influence of the parameter~$N_s$ on the performance of the proposed method, we repeated this experiment for different values of $N_s\in\{0,\ldots,8\}$ and evaluated the behavior of $\text{RMSE}_{\bA}$. The results, also seen in Table~\ref{tab:results_synthData}, show that $\text{RMSE}_{\bA}$ decreases with $N_s$. However, the performance improvements get small after about $N_s>3$, which indicates that a value of $N_s\leq3$ can yield a good compromise between abundance estimation \mbox{performance and computational complexity.}


\begin{figure}
    \centering
    \includegraphics[height=0.51\linewidth]{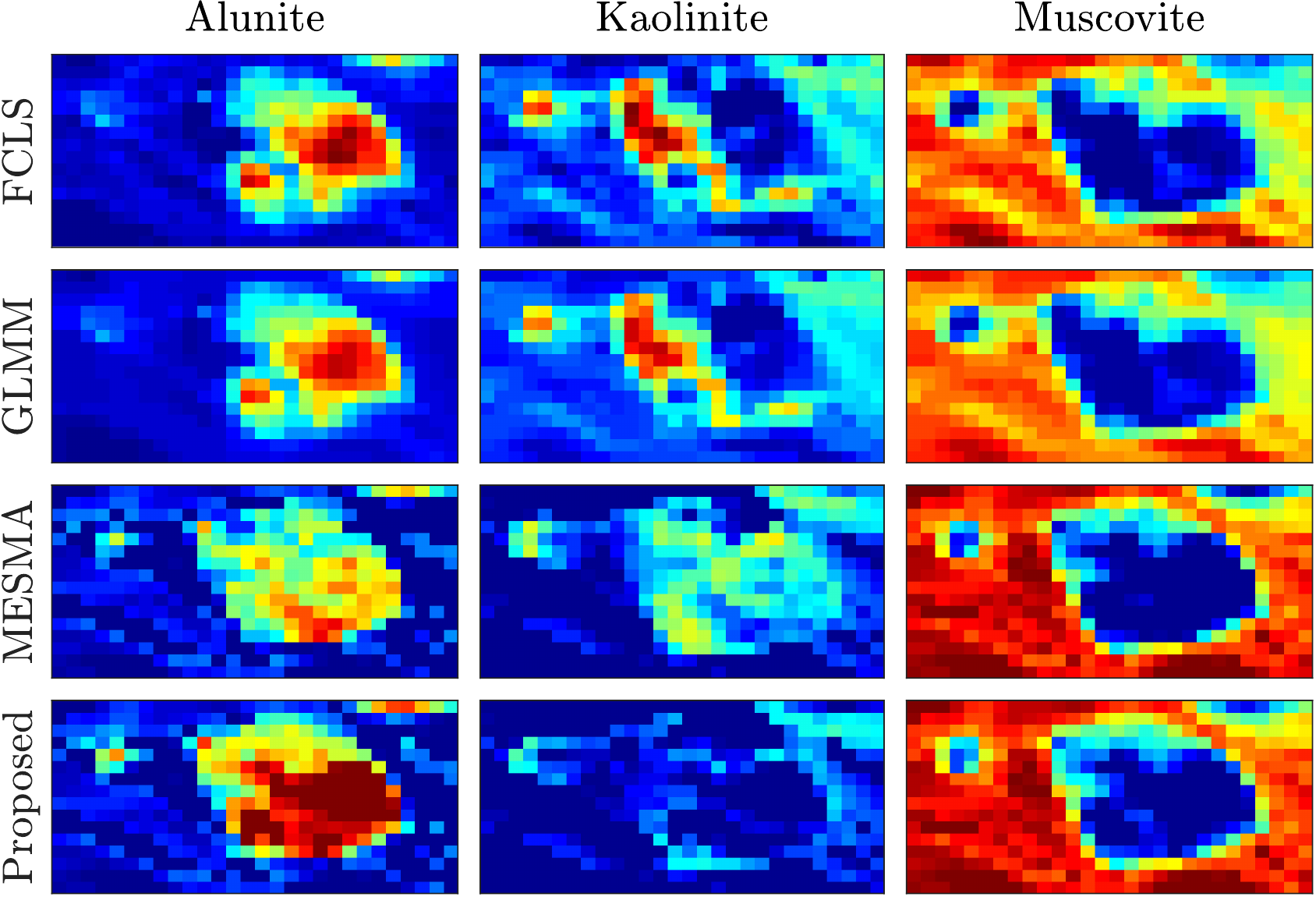}
    \includegraphics[height=0.51\linewidth]{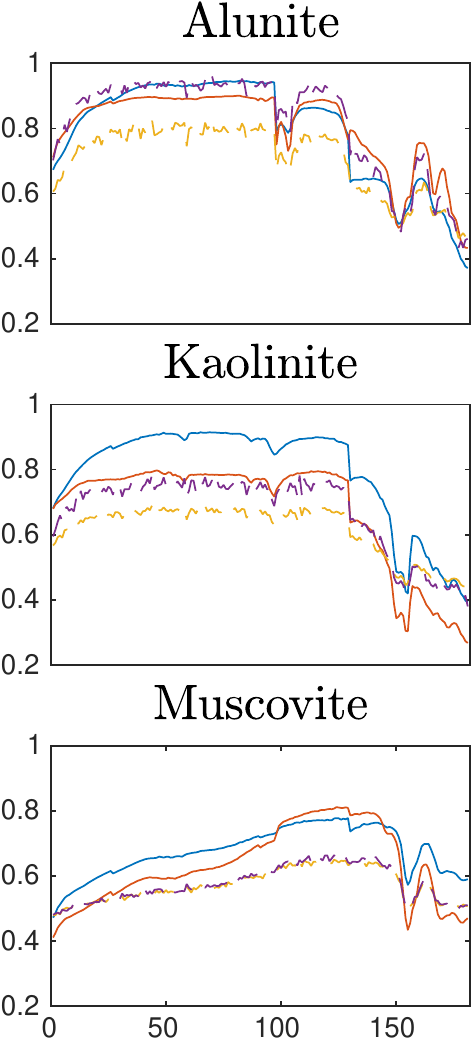}
    \vspace{-0.6cm}
    \caption{Left: abundance maps for the Alunite Hill subscene. Right: original endmembers (solid line) and synthetically generated signatures (dashed line).}
    \label{fig:results_alunHill_ab_and_em}
    \vspace{1.0ex}
    \includegraphics[width=0.75\linewidth]{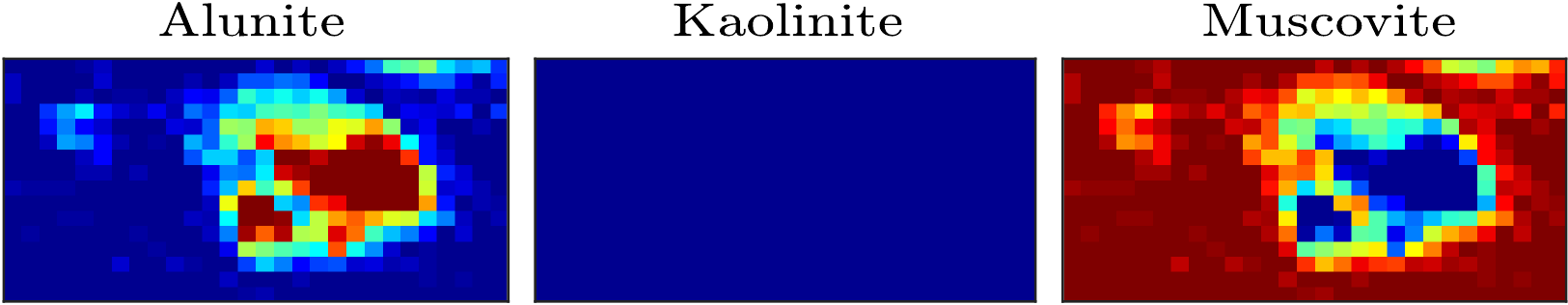}
    \vspace{-0.3cm}
    \caption{Abundance maps of MESMA for the Alunite Hill HI with a spectral library augmented using the Hapke model with known acquisition conditions.}
    \label{fig:aluniteHill_ab_HapkeLib}
    \vspace{-1.5ex}
\end{figure}



\paragraph*{\textbf{Real data}}

For the simulations with real data, we considered the Alunite Hill subscene of the Cuprite HI with $16\times28$ pixels, and a subscene of the Gulfport HI with $54\times70$ pixels~\cite{heylen2016alternatingAngleMinimization}. Water absorption or low SNR bands were removed and both the images and the spectral libraries were rescaled to have the same number of bands, resulting in $L=181$ for the Alunite Hill and $L=192$ for Gulfport.
These images were selected since the unmixing results can be evaluated using high-resolution classification maps available a priori, shown in Fig.~\ref{fig:aluniteHillGT}. The libraries $\mathcal{M}$ were built by selecting two signatures of each endmember from the USGS library and from field surveys (for the Alunite Hill and Gulfport HIs, resp.) such that the MESMA results closely approached (visually) the ground truth. $N_s=2$ additional signatures per EM were generated.

\begin{figure}[!t]
    \centering
    \includegraphics[height=0.48\linewidth]{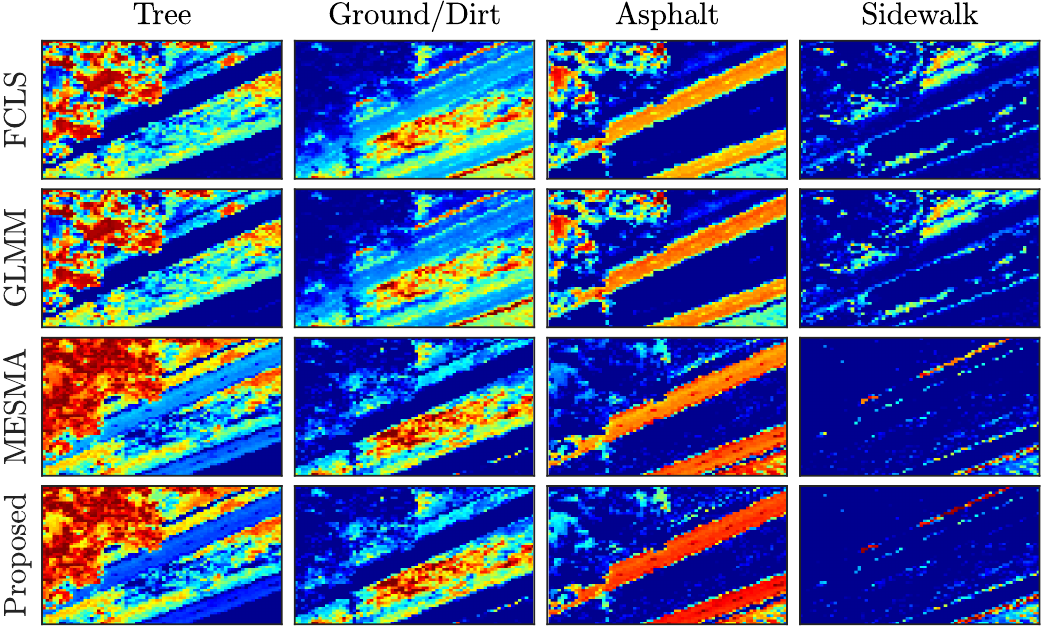}
    \includegraphics[height=0.48\linewidth]{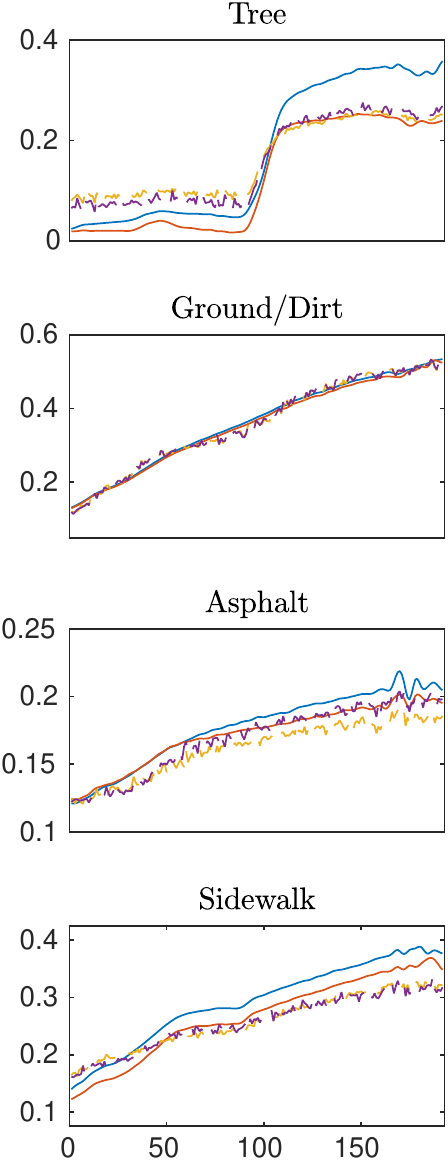}
    \vspace{-0.6cm}
    \caption{Left: abundance maps for the Gulfport subscene. Right: original endmembers (solid line) and synthetically generated signatures (dashed line).}
    \label{fig:results_gulfport_ab_and_em}
    \vspace{-1.5ex}
\end{figure}

The abundance maps reconstructed by all algorithms are provided in Figs.~\ref{fig:results_alunHill_ab_and_em} and~\ref{fig:results_gulfport_ab_and_em}. It can be seen that the abundance maps of the MESMA-based methods are significantly closer to the ground truth when compared to the GLMM and FCLS results. Furthermore, the proposed library augmentation strategy led to a much better representation of the alunite and kaolinite endmembers when compared to the competing approaches in the Alunite Hill HI. Similar results were obtained for the Gulfport HI, where the abundances obtained by the proposed method for the sidewalk and asphalt EMs approach the ground truth more closely when compared to those estimated by FCLS, GLMM and by MESMA with the original library.
The spectral signatures generated using the DGMs, also seen in Figs.~\ref{fig:results_alunHill_ab_and_em} and~\ref{fig:results_gulfport_ab_and_em}, show that the proposed strategy is able to generate signatures that accommodate variability seen in typical scenes from its representation in the original library. Specifically, a generally agreeable shape but different scaling variations that act nonuniformly over the spectral space can be seen in all cases except for the ground/dirt EM in the Gulfport HI, whose original field surveyed spectra (contained in $\mathcal{M}$) did not contain a meaningful amount of spectral variability.

The quantitative $\text{RMSE}_{\bY}$ results in Table~\ref{tab:results_synthData} show that the FCLS achieves smaller reconstruction errors in the real datasets when compared to the synthetic one, which contains more heavily mixed pixels and thus results in a worse data fitting for the FCLS (which is based on the VCA). However, we note that $\text{RMSE}_{\bY}$ is not a good measure of unmixing performance, as an infinite number of combinations (endmembers, abundances) often leads to the same reconstructed HI.

To compare the proposed method with physics-based library augmentation, we considered a Lambertian scattering approximation of the Hapke model to augment the library used with the Alunite Hill HI. Given prior knowledge about the laboratory acquisition conditions of the spectra in the USGS library, we can generate different variations of these mineral spectra by considering different viewing geometries as detailed in~\cite{heylen2014review}. The abundances estimated by MESMA using the augmented library are shown in Fig.~\ref{fig:aluniteHill_ab_HapkeLib}. Although a clear improvement can be seen in the alunite and muscovite EMs when compared to the original library, the kaolinite abundances were completely absorbed into the muscovite abundance map. Moreover, the alunite region is smaller than what is indicated in the ground truth, which is more closely matched by the results obtained using the proposed method. This shows that the proposed strategy can be competitive with physics-based models in practice.

\vspace{-1ex}

\section{Conclusions} \label{sec:conclusions}

In this work, a novel spectral library augmentation strategy was proposed for MESMA-like algorithms. Using the spectral signatures present in existing libraries as training samples, we applied deep generative models to learn the statistical distribution of endmember spectra. This allowed us to sample new spectral signatures from the estimated endmember distribution, which were then included in the augmented library, improving its ability to properly represent the materials present in practical scenes. Simulation results with both synthetic and real data showed that the proposed methodology can significantly improve the performance of the MESMA algorithm.


\bibliographystyle{IEEEtran}
\bibliography{abbrev_EMlibInterpolation}

\end{document}